%% file: main.tex
\documentclass[10pt,twocolumn,letterpaper]{article}

\usepackage{x}
\makeatletter
\@namedef{ver@everyshi.sty}{}
\makeatother
\usepackage{times}
\usepackage{epsfig}
\usepackage{graphicx}
\usepackage{booktabs}
\usepackage{amsmath}
\usepackage{amssymb}
\usepackage{multirow}
\usepackage{amsfonts}
\usepackage{amsmath}
\usepackage{subcaption}
\usepackage{xcolor}
\usepackage{makecell}
\usepackage{enumitem}

\usepackage[textwidth=1.8cm]{todonotes}
\pdfoutput=1

\DeclareMathOperator*{\argmax}{arg\,max}

\makeatletter
\newcommand\footnoteref[1]{\protected@xdef\@thefnmark{\ref{#1}}\@footnotemark}
\makeatother

\newcommand{\add}[1]{\fontsize{7}{4}\selectfont{\color{blue}{(+#1)}}}
\newcommand{\sub}[1]{\fontsize{7}{4}\selectfont{\color{red}{(-#1)}}}

\usepackage[pagebackref=true,breaklinks=true,letterpaper=true,colorlinks,bookmarks=false]{hyperref}

\finalcopy 

\def\xPaperID{1831} 

\ifxfinal\fi

\begin{document}

\title{Missing Modality Robustness in Semi-Supervised Multi-Modal \\Semantic Segmentation}

\author{Harsh Maheshwari\thanks{\href{mailto:harshmaheshwari@gatech.edu}{harshmaheshwari@gatech.edu}}
\qquad
Yen-Cheng Liu
\qquad
Zsolt Kira\\
Georgia Institute of Technology\\
}

\maketitle
\ifxfinal\thispagestyle{empty}\fi

\begin{abstract}
   Using multiple spatial modalities has been proven helpful in improving semantic segmentation performance.
    However, there are several real-world challenges that have yet to be addressed: (a) improving label efficiency and (b) enhancing robustness in realistic scenarios where modalities are missing at the test time. To address these challenges, we first propose a simple yet efficient multi-modal fusion mechanism \emph{Linear Fusion}, that performs better than the state-of-the-art multi-modal models even with limited supervision. Second, we propose \emph{M3L}: \emph{M}ulti-modal Teacher for \emph{M}asked \emph{M}odality \emph{L}earning, a semi-supervised framework that not only improves the multi-modal performance but also makes the model robust to the realistic missing modality scenario using unlabeled data. We create the first benchmark for semi-supervised multi-modal semantic segmentation and also report the robustness to missing modalities. Our proposal shows an absolute improvement of up to $10\%$ on robust mIoU above the most competitive baselines. Our code is available at \href{https://github.com/harshm121/M3L}{https://github.com/harshm121/M3L}
\end{abstract}

\input{sections/introduction.tex}

\input{sections/related.tex}
\input{sections/method.tex}
\input{sections/experiments.tex}
\input{sections/conclusion}
{\small
\bibliographystyle{ieee_fullname}
\bibliography{egbib}
}
\clearpage
\appendix
\input{sections/appendix}
\end{document}

%% file: sections/introduction.tex
\section{Introduction}

The availability of multiple sensors such as RGB, depth, and infrared has encouraged the use of multiple modalities for scene understanding tasks like semantic segmentation \cite{tokenfusion, cen, cfn, rdfnet, ssma, asymfusion, shapeconv}. Multi-modal semantic segmentation has shown promising results outperforming their uni-modal counterparts \cite{segformer, lin2017refinenet, chen2017rethinking} due to the effective use of auxiliary information present across modalities. However, one challenge with semantic segmentation is getting substantial amounts of annotated data, which is a  laborious and costly process. This has encouraged a need to create algorithms that work well with limited supervision. One approach is to utilize a mixture of labeled and unlabeled data (\textit{i.e.,} semi-supervised learning), and several works have approached this for various tasks \cite{liu2021unbiased, cai2022semi, weng2022semi, Liu_2022_CVPR}. To the best of our knowledge, however, all of the semi-supervised semantic segmentation research has been focused on uni-modal segmentation \cite{cct, gct, chen2021-CPS, u2pl, pc2seg, zou2020pseudoseg}. 
Thus, there is a need to explore semi-supervised frameworks for multi-modal semantic segmentation that can effectively use additional modalities to make the task label efficient.

\input{figures/teaser}
We find that there are two major challenges for making multi-modal segmentation models more useful that need attention. The first is to create a modality fusion algorithm that can work well even with limited supervision. 
The current multi-modal literature \cite{tokenfusion, cmx} has focused on fully supervised scenarios and thus the resulting methods do not necessarily work well with limited supervision. There has also been a growing interest in using transformers for multiple modalities due to their flexibility to incorporate various data types. The state-of-the-art multi-modal semantic segmentation method  \cite{tokenfusion} uses a learned fusion mechanism with a transformer-based segmentation architecture, Segformer \cite{segformer}. The use of transformers and training additional parameters to learn the fusion mechanism have made it more challenging for the existing models to perform well in a low-label regime. 

The second challenge is a lack of robustness to test-time missing modalities.  
Multi-modal models show an improvement over their uni-modal counterparts by effectively fusing the auxiliary information from different modalities. However, this improvement comes with a stricter requirement of guaranteeing that all of the modalities will be present during test time. This requirement could be difficult to satisfy due to sensor failures and unreliability. As previously discussed in the medical domain \cite{shen2019brain, urn, dorent2019hetero}, we also discover a major weakness with current multi-modal semantic segmentation - missing modality robustness. We find that if any modality is missing during test time, the segmentation performance of such models falls drastically, even below their uni-modal counterparts as depicted in Figure \ref{fig:teaser-figure}. 
To address the above limitations, we introduce \emph{Linear Fusion}, a multi-modal segmentation model that works effectively even with limited supervision, and \emph{M3L}: \emph{M}ulti-modal Teacher for \emph{M}asked \emph{M}odality \emph{L}earning, a semi-supervised framework which effectively uses unlabeled data to not only improve multi-modal semantic segmentation performance but also to make the model robust to missing modalities.

Specifically, \emph{Linear Fusion} is a simple yet effective multi-modal segmentation model that combines tokens from the two modalities linearly and thus learns cross-modal interaction without using additional trainable parameters. This makes the simple algorithm effective even when trained with limited supervision. 
For example, when trained with $0.2\%$ data on Stanford Indoor dataset \cite{stanfordindoor}, Linear Fusion outperforms the current state-of-the-art by $3.5\%$ points mIoU. In addition, to both leverage unlabeled data and enhance the robustness to missing modality, we propose \emph{M3L}, a semi-supervised framework that trains a Linear Fusion model and uses a multi-modal mean teacher to supervise a student network with a randomly chosen modality masked in the input. This makes the model robust to missing modalities while improving segmentation performance. Surprisingly, we find that a bi-modal model trained with our framework, when given a single input, still performs better than its uni-modal semi-supervised counterparts, and thus M3L can also be used to improve uni-modal semi-supervised semantic segmentation by using privileged multiple modalities during training (details in Section~\ref{sec:results-uni-semi-sup}). 

We perform extensive experimentation and comparison of our proposed methods against existing baselines to verify our claims and show the effectiveness of our proposals. We show that Linear Fusion, when trained with M3L, shows an improvement of up to $10\%$ points mIoU above the most competitive baseline on robust multi-modal segmentation. Moreover, for uni-modal segmentation, our method shows an absolute improvement of up to $3.5\%$ mIoU for RGB uni-modal and up to $6.5\%$ mIoU for depth uni-modal over the semi-supervised uni-modal segmentation baselines.\\
Finally, we list all our contributions:

\begin{enumerate}

\item To the best of our knowledge, we are the \textit{first} to address semi-supervised multi-modal semantic segmentation, and we create a new benchmark and evaluate robustness to realistic test-time missing modality scenarios.

\item We present a simple yet effective cross-modal integration mechanism, Linear Fusion, which outperforms state-of-the-art \cite{tokenfusion} under limited supervision.

\item We propose a semi-supervised training framework, M3L, which utilizes unlabeled images to improve the segmentation performance and make the model robust to missing modalities. 

\end{enumerate}

\noindent We release code and a demo to encourage further research.\footnote{\href{https://github.com/harshm121/M3L}{https://github.com/harshm121/M3L}, ~\href{https://harshm121-m3l.hf.space/}{https://harshm121-m3l.hf.space/}}.

%% file: figures/teaser.tex
\begin{figure}
\begin{center}
\includegraphics[width=\linewidth]{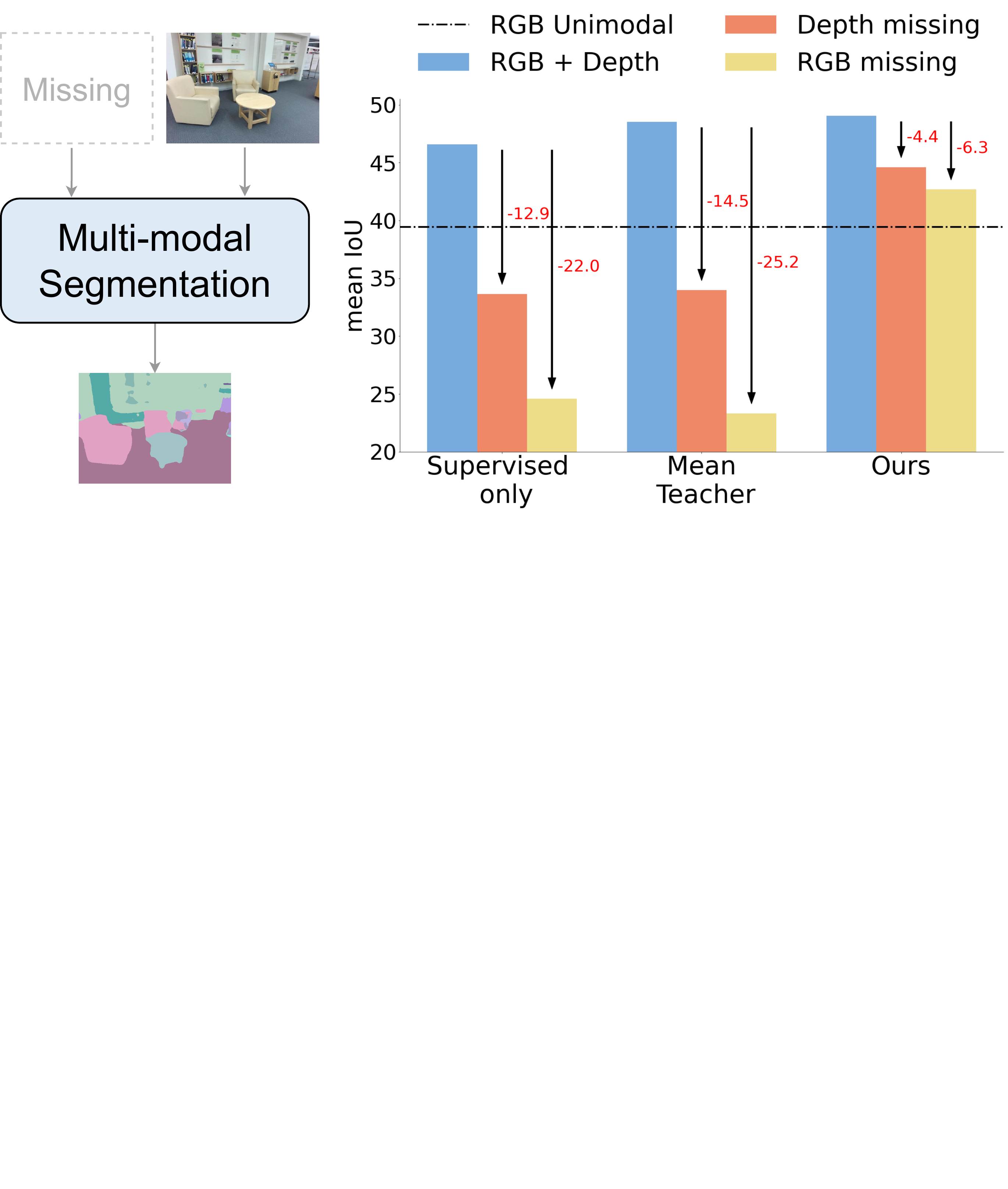}
\vspace{-6.5cm}
\end{center}
\caption{
Existing multi-modal semantic segmentation models require a significant amount of training data and exhibit lower robustness when dealing with missing modalities during test time. Our proposed framework can learn the model with a limited amount of labeled data and enhance the robustness to test-time missing modality scenarios.
}
\label{fig:teaser-figure}
\vspace{-3mm}
\end{figure}

%% file: sections/related.tex
\section{Related Work}

\input{tables/related}

\noindent\textbf{Multi-modal semantic segmentation.} As accessing multiple spatial modalities like RGB, depth, and infrared is getting easier, many methods have been proposed to use more than one modality to improve semantic segmentation performance. The holy grail of the multi-modal semantic segmentation community is to find effective ways to fuse auxiliary information from multiple modalities. These methods can be broadly categorized into early \cite{deng2019rfbnet, hu2019acnet, sun2019rtfnet}, late \cite{zhang2019exploration}, and mid/hybrid \cite{rdfnet, cen, tokenfusion, ssma, jiang2018rednet, cmx} fusion techniques. 
Convolutional models have dominated this literature so far, and interest has arisen in using transformer-based architectures for fusing multiple modalities \cite{tokenfusion, cmx}. 
However, these works focused on creating fusion mechanisms for fully supervised multi-modal semantic segmentation, and there is no prior work on \textit{multi}-modal semi-supervised semantic segmentation. 

\noindent\textbf{Semi-supervised semantic segmentation.} Creating labels for segmentation is a more laborious task and can cost around $25\times$ more than getting labels for classification\footnote{\href{https://cloud.google.com/ai-platform/data-labeling/pricing}{https://cloud.google.com/ai-platform/data-labeling/pricing}}. This has motivated a lot of research on semi-supervised semantic segmentation \cite{chen2021-CPS, u2pl, zou2020pseudoseg, pc2seg, gct, cct, liu2022perturbed}. We find that most of these methods are smart extensions of the popular mean teacher framework \cite{mt} which uses a weight-ensembled teacher model to generate pseudo labels. However, we found that  all of the semi-supervised semantic segmentation work focused on only RGB uni-modal models, and an investigation of how additional modalities can be used to more strongly leverage unlabeled data is an interesting research question. 

Thus, we found a rising interest in semi-supervised learning using multiple modalities in other domains such as medical \cite{chartsias2020disentangle, 9250615}, videos \cite{xiong2021multiview}, speech \cite{sunkara2020multimodal} but not for semantic segmentation. Our proposed approach thus uses a semi-supervised setup for multi-modal semantic segmentation and extends the mean teacher \cite{mt} framework to not only improve segmentation performance but also make the model robust to missing modalities.

\noindent\textbf{Missing modality robustness.} The use of multiple modalities is common in various domains and thus the robustness to missing modalities has become an important area of focus. For example, in medical domains, it is possible to not have access to all modalities (all types of scans like MRI, CT, etc.) for all patients. 
Thus missing modality robustness in the medical domain has caught a lot of attention.
The approaches mainly include either synthesizing the missing modality \cite{jog2017random, sharma2019missing, yu20183d}, or learning a shared latent space for all modalities \cite{shen2019brain, urn, dorent2019hetero} or knowledge distillation methods \cite{wang2020multimodal, azad2022smu}. Modality dropout has also been used to improve performance in certain domains by ensuring enough attention to all modalities and as a regularization to not let the dominant modality drive the prediction. 
Neverova \textit{et al.}~\cite{neverova2015moddrop} proposed ModDrop as augmentation for gesture recognition, Abdelaziz \textit{et al.}~\cite{hussen2020modality} proposed modality dropout for driving animated talking faces for audio-video modalities and \cite{9746613} for multi-modal dialogue systems. 
We show that just using the modality dropout augmentations is not enough for a semi-supervised training framework and thus M3L uses a knowledge distillation framework to utilize the unlabeled data to make the models robust to missing modality.
Frameworks that gain performance by utilizing multiple modalities also become prone to failure when the modalities are not guaranteed to be present during test time. This has gained attention in other domains and we highlight the same for multi-modal semantic segmentation. 

\noindent\textbf{Robustness in segmentation.} In this work, we focus on robustness to missing modalities at test time. Robustness in segmentation also can be to other forms of degradation. Tian \textit{et al.}~\cite{tian2020uno} presents a method to fuse multiple modalities for segmentation effectively when certain modalities may suffer from degradations like  motion blur, gaussian noise, fog, etc. Robustness to seasonal changes or lighting effects has also been discussed in prior works~\cite{Larsson_2019_CVPR, valada2016towards, 9011192}. However, our work focuses on a scenario when the entire modality is missing, which is another possible situation due to sensor malfunction or other unreliabilities.

We summarise the related work in Table~\ref{tab:related-work}. We focus on a problem setting that has been explored in parts. There has been a growing interest in multi-modal segmentation and semi-supervised segmentation. We also find a growing interest in missing modality robustness in other domains but it has not been discussed for semantic segmentation. Thus, to make segmentation models more useful and practical, there is a need to address both, the label-efficiency and the robustness to missing modalities of such models. We thus propose our method M3L to address both challenges.

%% file: tables/related.tex
\begin{table*}[ht]
\centering
\resizebox{0.85\linewidth}{!}{%
\begin{tabular}{ccccc}
\toprule
Related Area & Multi-modal & Semi-supervised & Missing modality robustness & Related work\\ \midrule
Multi-modal segmentation & \checkmark &  &  & \cite{deng2019rfbnet, hu2019acnet, sun2019rtfnet, zhang2019exploration, rdfnet, cen, tokenfusion, ssma, jiang2018rednet, cmx} \\
Semi-supervised segmentation & & \checkmark &  & \cite{chen2021-CPS, u2pl, zou2020pseudoseg, pc2seg, gct, cct, liu2022perturbed} \\
Missing modality robustness & \checkmark && \checkmark & \cite{jog2017random, sharma2019missing, yu20183d, shen2019brain, urn, dorent2019hetero, wang2020multimodal, azad2022smu, hussen2020modality, 9746613} \\
Ours & \checkmark & \checkmark & \checkmark  & - \\ \bottomrule
\end{tabular}
}
\caption{Related literature. Missing modality robustness is unexplored for semantic segmentation. We address the two simultaneous challenges of making the multi-modal models robust to missing modality while making them label efficient, thus more practical and useful.
}
\label{tab:related-work}
\end{table*}

%% file: sections/method.tex
\section{Method}

\noindent\textbf{Problem definition.} Our ultimate goal is to address the missing modality robustness of multi-modal segmentation models when trained with limited supervision. We consider our data has two modalities for training: RGB, denoted by $x^{rgb}$, and depth, denoted by $x^{depth}$. The goal is to output a segmentation map classifying each pixel into one of $\mathbb{C}$ classes.  We consider a set of labeled samples, $\mathcal{D}_s = \{x_i^{rgb}, x_i^{depth}, y_i\}_{i=1}^{N_s}$ and a set of unlabeled samples $\mathcal{D}_u = \{x_i^{rgb}, x_i^{depth}\}_{i=1}^{N_u}$. $N_s$ and $N_u$ are the numbers of labeled and unlabeled data.
To examine the performance and robustness of any algorithm, say $\mathcal{A}$, we report the performance $P$ of the predictions under three test conditions: a) Using both the modalities: $P(\mathcal{A}(x^{rgb}, x^{depth}), y)$, b) with RGB only: $P(\mathcal{A}(x^{rgb}), y)$ and c) with depth only: 
$P(\mathcal{A}(x^{depth}), y)$.

To address missing modality robustness, we propose M3L, a semi-supervised training framework for making the multi-modal semantic segmentation models robust to missing modalities (Section \ref{sec:method-M3L}). We do so by devising a knowledge distillation framework, leveraging pseudo labels from a multi-modal teacher to a masked modality student model. However, before talking about the semi-supervised framework, we pay attention to the base multi-modal model and find that the existing state-of-the-art fusion mechanism~\cite{tokenfusion} does not perform well in a low-label regime (Section \ref{sec:revisiting-mm}). Hence, we first devise Linear Fusion, a simple yet effective fusion mechanism for multi-modal semantic segmentation (Section \ref{sec:method-lf}), and then train the Linear Fusion model using M3L, to improve the segmentation performance and make the model robust to missing modalities.

\subsection{Revisiting Multi-modal Semantic Segmentation}
\label{sec:revisiting-mm}

With the goal of improving semantic segmentation using auxiliary information from different spatial modalities, Wang et al. proposed TokenFusion~\cite{tokenfusion}. It is a multi-modal transformer architecture that dynamically detects uninformative tokens from a modality and substitutes them with tokens from other modalities allowing the transformer to learn cross-modal interactions. 
The detection of uninformative tokens is achieved by thresholding a score estimated by a separate scoring module. 
Unlike Segformer~\cite{segformer} (Eq. \ref{eq:normal-transformer}), TokenFusion (Eq. \ref{eq:token-fusion-scoring}) uses the score from the scoring module as weights for each token before passing it to the next layer. The substitution is done as shown in Eq. \ref{eq:token-fusion-exchange} and is encouraged by making the scores sparse using $\mathbf{L1}$ loss on the scores to learn cross-modal interactions.
\begin{equation}
    \hat{e}_{m}^{l} = \text{MHA}(\text{LN}(e_m^l)), e_{m}^{l+1} = \text{FF}(\text{LN}(\hat{e}_{m}^{l}))
    \label{eq:normal-transformer}
\end{equation}
\begin{equation}
    \hat{e}_{m}^{l} = \text{MHA}\big(\text{LN}(e_m^l) \cdot s^l(e^l_m)\big), e_{m}^{l+1} = \text{FF}(\text{LN}(\hat{e}_{m}^{l}))
    \label{eq:token-fusion-scoring}
\end{equation}
\begin{equation}
    e_m^l = e_m^l \odot \mathbb{I}_{s^l(e^l_m) \geq \theta} + e_{m'}^l \odot \mathbb{I}_{s^l(e^l_m) < \theta}
    \label{eq:token-fusion-exchange}
\end{equation}
Here, $e_m^l$ denotes the tokens out of the $l^{th}$ layer for the $m^{th}$ modality, and $\theta$ is the exchange threshold. MHA, LN, and FF are Multi-Headed Attention, Layer Normalization, and Feed Forward modules as described in Segformer \cite{segformer}.

\noindent\textbf{Limitation.} We find that this learned substitution mechanism does not work very well (more details in Section~\ref{sec:results-lf}) when trained with small amounts of data.
To this end, we propose a much simpler fusion mechanism that has significant performance improvement over Token Fusion.

\subsection{Linear Fusion for Cross-Modal Integration}
\label{sec:method-lf}
Fusing information from multiple modalities is the holy grail of multi-modal segmentation. 
Multiple methods have been presented in the literature for convolution models \cite{cen, cfn, rdfnet, ssma, asymfusion} but only recently has the interest arisen to use transformer models for using multiple modalities for segmentation. 
However, it is a challenging task with limited supervision where there is not enough data to train complex learning-based fusion mechanisms. 

\input{figures/overview_linearfusion}

\noindent\textbf{Base Model.} We consider Segformer \cite{segformer}, as our base segmentation model based on the transformer architecture due to its popularity in the vision community.
To extend the uni-modal Segformer architecture to handle multiple input modalities, we create two copies of the hierarchical transformer encoder $f$ from Segformer which share weights $\theta$. 
However, as proposed in Token Fusion~\cite{tokenfusion}, we use separate layer normalization parameters $\gamma$ for the two modalities as the statistics of different modalities can be vastly different. 
Each branch gets a single modality $x^m$ as input and the final representation of each branch is passed to a lightweight MLP decoder $g$ with parameters $\phi$, as proposed in Segformer. Specifically, 

\begin{equation}
\hat{y}^m = g_\phi \circ f_{\theta, \gamma^m}(x^m, \mathbf{e}^{\overline{m}}),
\label{eq:encoder-decoder-lf}
\end{equation}
 where $\hat{y}^m$ is the prediction from a branch corresponding to the modality $m$. Note that the encoder $f$ takes $\mathbf{e}^{\overline{m}}$ as an additional argument, which is a vector of all the tokens from intermediate layers of the other branch $\overline{m}$ and is used for fusion, as described below.

\input{figures/overview-M3L}

\noindent\textbf{Fusion.} 
We apply a simple fusion mechanism that integrates cross-modal tokens by using a linear combination.

 Specifically, each of the two branches gets a single modality $x^m$ as input to the encoder $f$. The encoder $f$ has $L$ attention layers and the intermediate token outputs from each attention layer $e_l$ are fused together, 

\begin{equation}
     e_m^l = \alpha \times e_m^l + (1-\alpha) \times e_{\overline{m}}^l,
    \label{eq:fusion-lf}
\end{equation} 
where the hyperparameter $\alpha$ is the fusion weight guiding the extent of information preserved in the branch. 
In this way, each branch combines the information it receives from the other branch by linearly combining the tokens. We depict this fusion mechanism in Figure~\ref{fig:linear-fusion}. $\mathbf{e}^{m}$ represented in Eq. \ref{eq:encoder-decoder-lf} is a vector of $[e_m^l]_{l=1}^{L}$. 

\noindent\textbf{Ensemble prediction.} As shown in Eq. \ref{eq:encoder-decoder-lf}, each branch outputs an individual prediction driven by input from a modality and fusion of information from the other. The final prediction of the model is a weighted ensemble of $\hat{y}^m ~\forall~ m$. 
\begin{equation}
    \hat{y} = \lambda \hat{y}^{rgb} + (1-\lambda)\hat{y}^{depth}
    \label{eq:ensemble-lf}
\end{equation}
where $\lambda$ is a trainable ensemble weight. Essentially, the model outputs three predictions $\hat{y}^{rgb}$, $\hat{y}^{depth}$ from the two branches and the ensemble $\hat{y}$. The model is trained using supervised loss $\mathcal{L}_s$ which is the average of segmentation loss $\mathcal{L}_{seg}$ of all three predictions with the ground truth $y$,
\begin{equation}
    \mathcal{L}_s = \frac{1}{3}\Big[\mathcal{L}_{seg}(\hat{y}, y) + \mathcal{L}_{seg}(\hat{y}^{rgb}, y) + \mathcal{L}_{seg}(\hat{y}^{depth}, y)\Big]
    \label{eq:loss-lf}
\end{equation}

The above method describes an effective way to integrate the cross-modal features, and our empirical results in Section~\ref{sec:results-lf} indicate our simple linear fusion performs favorably against the prior work~\cite{tokenfusion} under the low-label (and even the full-labeled)
settings.
With this base segmentation model, we now proceed towards a semi-supervised framework to help with missing modality robustness for multi-modal segmentation.

\subsection{M3L: Multi-modal Teacher for Masked Modality Learning}
\label{sec:method-M3L}
To further improve label efficiency, in this section we propose a training framework that leverages unlabeled data to improve the performance of the multi-modal segmentation model and makes the model robust to missing modalities. We consider the base segmentation model to be Linear Fusion and denote it by $LF$ which outputs a segmentation prediction $\hat{y} = LF(x^{rgb}, x^{depth})$.
As presented in Figure~\ref{fig:M3L}, we describe our M3L framework as follows.\\

\noindent\textbf{Supervised training.} For the labeled samples in $\mathcal{D}_s$, we compute the segmentation loss described in Eq. \ref{eq:loss-lf} to train the base segmentation model, as depicted in Figure \ref{fig:M3L} (a). \\

\noindent\textbf{Unsupervised training.} To leverage the unlabeled data for improving the performance and missing modality robustness, we propose a semi-supervised framework based on a teacher-student mechanism \cite{mt} and a modality dropout scheme as shown in Figure \ref{fig:M3L} (b). Specifically, the framework consists of a teacher ($LF^t$) and a student ($LF^s$) model which are identical network architectures but do not share weights\footnote{We overload the notation for simplicity and use $\theta_t$ and $\theta_s$ to represent the teacher and student's parameters.}.

Our teacher model, $LF^t$ takes both the modalities (RGB \textbf{and} depth) as input and estimates the segmentation mask using Eq. \ref{eq:encoder-decoder-lf} and \ref{eq:ensemble-lf}.

\begin{equation}
    \hat{y}_t = LF^t(x^{rgb}, x^{depth})
    \label{eq:M3L-teacher-pred}
\end{equation}

We then generate \emph{hard} pseudo-labels, $y_{p} = \argmax ~\hat{y}_t$. To improve the label-efficacy of the student model and to make it more robust to the missing modality, we propose to randomly mask 100\% 

of either modality (RGB \textbf{or} depth) in the input to the student model. 
To handle the missing modality, we use a learnable token to fill in for all the missing tokens of the masked modality.
The learned token can thus be used during inference whenever any modality is missing, as shown in Figure~\ref{fig:M3L} (c).
Modality dropout makes the student model robust to missing modalities by not only providing a learned token to fill-in whenever needed but also by encouraging the model to pay attention to all modalities and discouraging the dominating modality to overpower.

The student predictions are supervised using the \emph{hard} pseudo-labels generated by the teacher.

\small{
\begin{equation}
    \mathcal{L}_u = \frac{1}{3}\Big[\mathcal{L}_{useg}(\hat{y}_{s}, y_{p}) + \mathcal{L}_{useg}(\hat{y}_s^{rgb}, y_{p}) + \mathcal{L}_{useg}(\hat{y}_s^{depth}, y_{p})\Big]
    \label{eq:M3L-unsup-loss}
\end{equation}
}
where $\mathcal{L}_{useg}$ is the unsupervised segmentation loss. The loss $\mathcal{L}_u $ is computed on both the labeled and unlabeled samples.

\noindent\textbf{Overall loss.}
We train the overall framework using a batch of both labeled and unlabeled samples. The supervised loss $L_s$ (Eq. \ref{eq:loss-lf}) is computed on the labeled samples and the unsupervised loss $L_u$ (Eq. \ref{eq:M3L-unsup-loss}) is computed on both the labeled and unlabelled samples. 
\begin{equation}
    \mathcal{L} = \sum_{(\mathbf{x}, y)\sim \mathcal{D}_s} \mathcal{L}_s(\mathbf{x}, y) + \lambda_{pseudo}\sum_{(\mathbf{x}, y)\sim \mathcal{D}_s \cup \mathcal{D}_u} \mathcal{L}_u(\mathbf{x})
    \label{eq:M3L-total-loss}
\end{equation}

We train the student model by backpropagating the total loss $\mathcal{L}$ but detach the teacher parameters from the computation graph. The teacher model's parameters are updated slowly using the student model's parameters. 

\noindent\textbf{Teacher update.}
To obtain stable pseudo-labels from the teacher model, we set the teacher model's parameters as the exponential moving average (EMA) of the student's parameters. The slowly progressing teacher can be regarded as a temporal ensemble of the student model across training iterations.
\begin{equation}
    \theta_t \leftarrow \alpha_{ema}\theta_t + (1-\alpha_{ema})\theta_s
    \label{eq:M3L-teacher-update}
\end{equation}
with $\alpha_{ema}$ being the hyperparameter controlling the rate of update of the teacher parameters. EMA teacher proposed in \cite{mt} has been successfully used in various other tasks \cite{cai2022semi, liu2021unbiased, u2pl, liu2022perturbed}.

%% file: figures/overview_linearfusion.tex
\begin{figure}
\begin{center}
\includegraphics[width=0.85\linewidth]{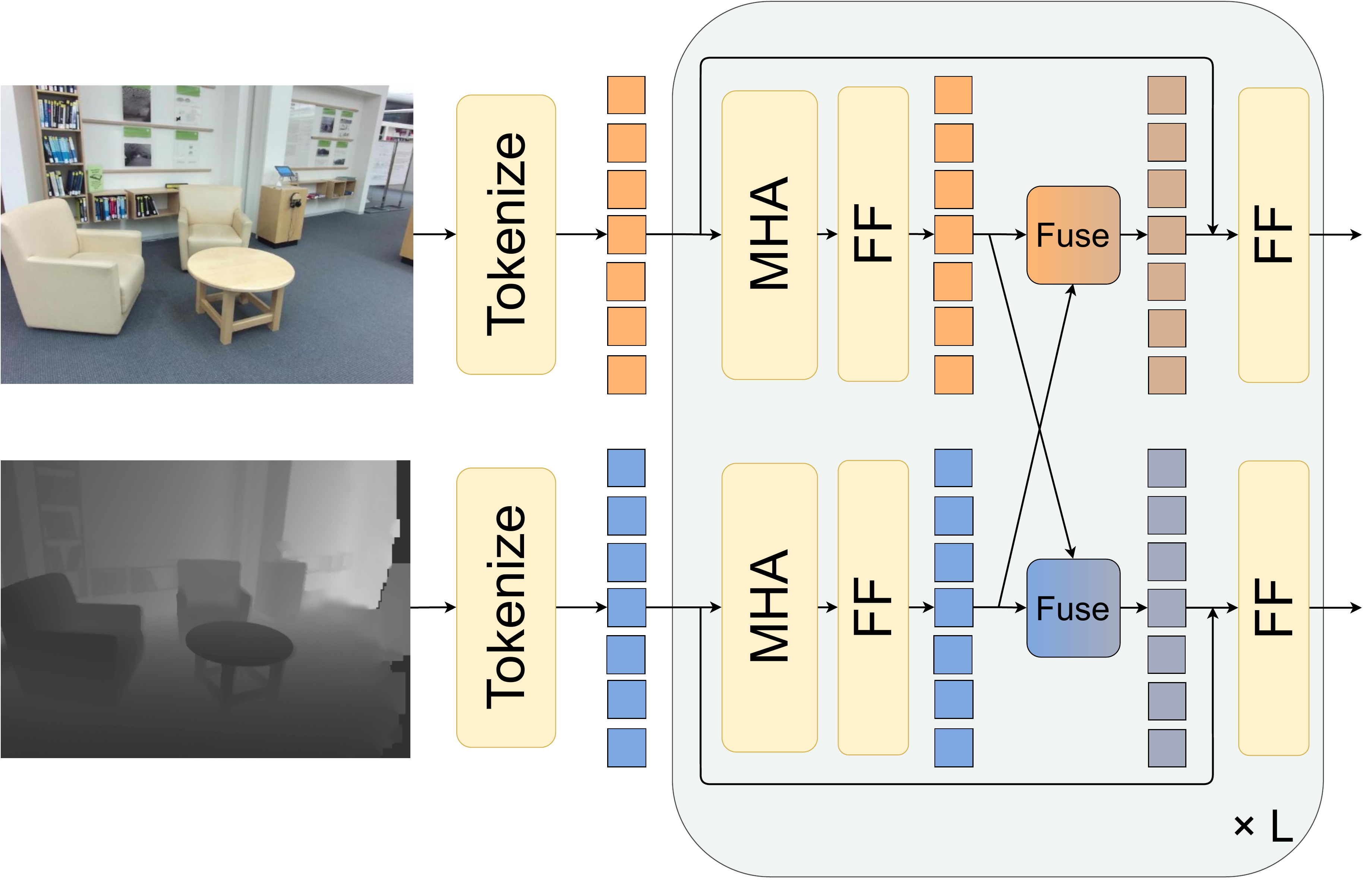}
\end{center}
\vspace{-5mm}
\caption{Overview of the Linear Fusion model. Information from the two modalities is fused by linearly adding the tokens from each branch. The fused tokens are then passed to further layers. MHA denotes Multi-Headed Attention and FF denotes a feed-forward module. Tokenize, MHA, and FF are the same as in the Segformer \cite{segformer} architecture}
\label{fig:linear-fusion}
\end{figure}

%% file: figures/overview-M3L.tex
\begin{figure*}
\begin{center}
\includegraphics[width=0.85\linewidth]{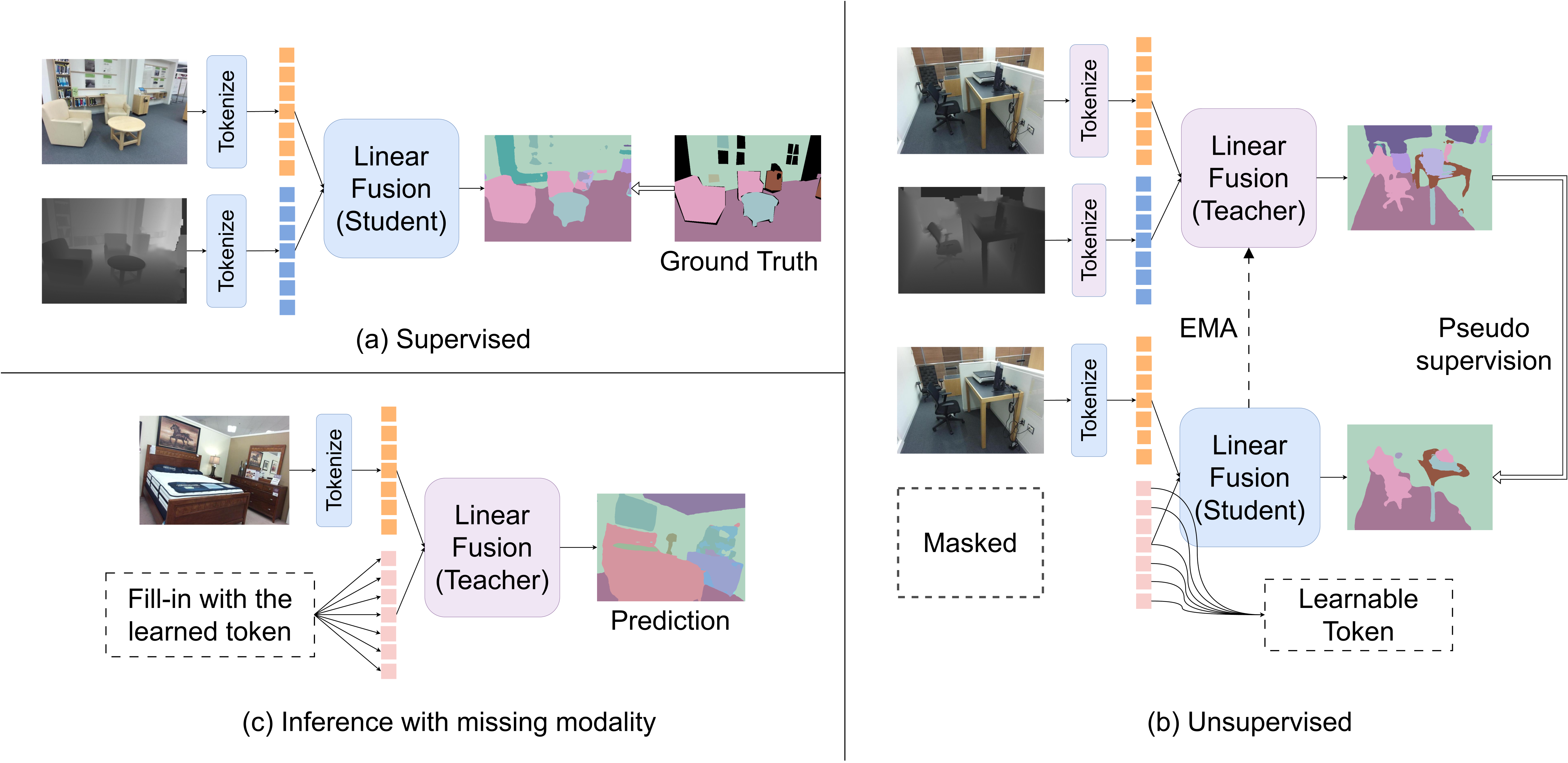}
\end{center}
\vspace{-5mm}
\caption{Overview of the M3L framework. (a) M3L supervises the prediction using ground truth for the labeled instances. (b) For the unsupervised loss, M3L uses a multi-modal mean (EMA) teacher which generates a segmentation prediction that is used to supervise a student. A randomly chosen modality is masked entirely in the student's input and a single learnable token is used to fill in the missing tokens. (c) The learned token thus can be used during inference if any modality is missing.
}
\label{fig:M3L}
\end{figure*}

%% file: sections/experiments.tex
\section{Experiments}

\input{tables/missing-modality-semi-sup-both}
\input{tables/uni-semi-sup}
\input{tables/results-lf}
\input{tables/ablation-M3L}
\input{tables/alphatune}

To demonstrate the effectiveness of Linear Fusion (Section~\ref{sec:method-lf}) and M3L (Section~\ref{sec:method-M3L}), we show empirical evidence by comparing against many baselines and ablate over our design choices.
\subsection{Benchmark for Semi-supervised Multi-modal Segmentation}
Since no prior work explores semi-supervised multi-modal semantic segmentation, we create a \textit{new} benchmark for this setting
\footnote{To the best of our knowledge, we are the first to create a benchmark for semi-supervised multi-modal semantic segmentation.}. We consider the challenging scenarios of missing modality (RGB-only or depth-only) and evaluate the robustness in the proposed and existing multi-modal semantic segmentation models.\newline

\noindent\textbf{Datasets.} Following prior multi-modal semantic segmentation literature \cite{tokenfusion, ssma}, we build the benchmark using indoor RGBD segmentation datasets.

We use two popular semantic segmentation datasets - Stanford Indoor \cite{stanfordindoor} and SUN RGBD \cite{sunrgbd}. 
We apply a more rigorous experiment setup, which splits the training into training and validation and keeps the testing set intact\footnote{We find the previous benchmark for multi-modal semantic segmentation~\cite{tokenfusion, ssma, cen, shapeconv} do not have validation set, which is possible to suffer from the overfitting issue on the test set.}.

We hope this encourages better ML practices by tuning the hyperparameter on only the validation set. 
We then create configurations of labeled/unlabeled training sets of varying sizes for semi-supervised training.

Stanford-Indoor \cite{stanfordindoor} is a large-scale dataset with $17593$ test samples and (originally) $52903$ train samples. 
All the samples correspond to 6 areas (areas 1-6) with area 5 used for the test set \cite{stanfordindoor}. 
We split the original $52903$ training samples corresponding to the other 5 areas into train/val sets of sizes ($49199$ / $3704$) with area 3 being used for validation. 
The dataset has 13 classes for the classification of pixels. 
We create three different semi-supervised configurations using $0.1\%$ ($49$), $0.2\%$ ($98$), and $1\%$ ($491$) of the data as labeled and the rest as unlabeled ($49159$, $49101$, $48708$).

SUN RGBD is a challenging dataset with $37$ classes and has $5049$ test samples. 
It originally had $5285$ training samples which we split into 90/10 train/val ($4757$ / $528$) samples. 
For semi-supervised training, we create three configurations and treat $6.25\%$ ($297$), $12.5\%$ ($594$), and $25\%$ ($1189$) of the data as labeled and the rest as unlabeled ($4460$, $4163$, $3568$). \\

\noindent\textbf{Baselines.} We compare our proposed M3L semi-supervised framework against supervised-only and mean teacher \cite{mt} to prove its efficacy. We also compare the performance of Linear Fusion with supervised-only transformer-based fusion approaches presented in the literature~\cite{tokenfusion, urn}. 
\begin{itemize}[leftmargin=*]
    \item URN~\cite{urn}: We choose a straightforward fusion mechanism that closely resembles the approach presented in Unified Representation Network~\cite{urn}, but we use Segformer as the base architecture. Different encoders are used for the two modalities and the encoded representations are fused (averaged) before passing them to a single decoder. Since two encoders are trained, URN has $\sim2\times$ the number of parameters than other methods described below.
    \item TF \cite{tokenfusion}: The state-of-the-art Token Fusion \cite{tokenfusion} framework. We train the model using our setup to report the performance.
    \item LF: Linear Fusion method as proposed in Section \ref{sec:method-lf}.
    \item LF + MT \cite{mt}: We train Linear Fusion with the semi-supervised mean-teacher \cite{mt} framework and report its performance. 
    \item LF + M3L: We train Linear Fusion with the proposed M3L semi-supervised framework as proposed in Section~\ref{sec:method-M3L}.
\end{itemize}

\noindent\textbf{Metrics.} To report the performance on the test set, we do \emph{single-scale}, \emph{non-sliding} testing by rescaling the input images to the expected model-input size and rescaling the predictions back to the original ground truth size using bilinear interpolation as done before \cite{chen2021-CPS, u2pl}. 
We report the mean IoU of all the methods when tested under 3 types of inputs: RGB+depth, RGB-only (depth missing), and depth-only (RGB missing), and then report the average of the three testing scenarios. We name this average performance as the \textit{MM-Robust} performance to quantify the missing modality robustness of multi-modal semantic segmentation. For M3L, we use the learned token during inference when a modality is missing.  For others, we give an all-zero input for the modalities that are missing. We also report the mean class accuracy and pixel accuracy for all results in the appendix.

\subsection{Implementation Details}
For all our models, we use the transformer-based model MiT-B2 proposed by Segformer \cite{segformer} for a fair comparison. We initialize the network with ImageNet-1k pre-trained checkpoint available publicly~\cite{segformer}.
For Linear Fusion, we tuned our fusion weight on the validation set and chose $\alpha = 0.8$ for all settings. 
For M3L, we either give RGB and Depth, or RGB-only, or Depth-only input to the student model (in equal proportions). We set $\alpha_{ema} = 0.99$ and $\lambda_{pseudo} = 1.0$. 
We use AdamW optimizer \cite{adamw} and train on a minibatch of 16 with a learning rate of $1e-4$ for the encoder and $3e-4$ for the decoder with a momentum of $0.9$, a weight decay of $1e-4$ and a polynomial decay of power $0.9$. 
For semi-supervised methods, we sample 16 labeled and 16 unlabeled data instances in each minibatch.
We use OHEM loss \cite{ohem} as $\mathcal{L}_{seg}$ in Eq. \ref{eq:loss-lf} and the multi-class cross-entropy loss as $\mathcal{L}_{useg}$ in Eq. \ref{eq:M3L-unsup-loss}. We train for $\sim15$k
iterations and pick the checkpoint (sampled after every $300$ iterations) with the best validation performance to report the test performance. Any hyperparameter tuning was done on the validation set keeping the test set untouched.
For training, we scale the images with a random factor between $[0.5, 2]$ and perform a random crop of $500\times500$ for SUN RGBD and $540\times540$ for Stanford Indoor. 
The code is implemented using PyTorch's Data Distributed Parallel and was run on 4 Nividia A40 GPUs. 

\subsection{Results}

To prove the efficacy of M3L and Linear Fusion, we perform extensive experiments on Stanford Indoor and SUN RGBD datasets and three labeled/unlabeled configurations per dataset and compare the performance against state-of-the-art baselines. \vspace{-0.5cm}
\subsubsection{Semi-supervised Multi-modal Semantic Segmenation}
\label{sec:results-missing-mod-semi-sup}

To show that M3L is effectively using unlabeled images, we compare the Linear Fusion model trained with M3L against the supervised-only baseline and a competitive mean teacher \cite{mt} baseline. When tested with multi-modal RGBD input, our proposed method consistently performs better and gives an absolute improvement of up to $2.01\%$ mIoU over the strong MT \cite{mt} baseline on Stanford Indoor \cite{stanfordindoor} dataset, as shown in Table \ref{tab:missing-modality-semi-sup-both} (a).

To show M3L’s ability to make the model robust to missing modalities, we also test the models on a more challenging scenario of missing modalities and report the MM-Robust metric, which is the average of the three possible test-time scenarios (missing depth, missing RGB, RGBD). As shown in Table \ref{tab:missing-modality-semi-sup-both} (a), on MM-Robust metric, M3L is better than the MT baseline by up to $9.99\%$ mIoU and consistently shows improvement for both RGB (depth missing) and depth (RGB missing) scenarios on Stanford Indoor dataset. We show similar results on the SUN RGBD dataset in Table \ref{tab:missing-modality-semi-sup-both} (b) and show an improvement of up to $4.05\%$ mIoU as measured by MM-Robust. Thus, M3L not only improves the multi-modal segmentation performance but also makes the model robust to missing modalities by effectively using the unlabeled data.

We note that on SUN RGBD dataset, even though M3L successfully improves the performance for missing modality scenarios, neither MT \cite{mt} nor M3L sufficiently improves the multi-modal (RGBD) performance. We attribute this to a lack of a large unlabeled set, which is even more essential for a challenging dataset like SUN RGBD with 37 fine-grained classes. 

\subsubsection{Comparison to Uni-modal Models}

\label{sec:results-uni-semi-sup}

To show M3L's uni-modal performance, we train Linear Fusion with the M3L framework utilizing both the modalities during training and test with only a single modality at test-time and compare against semi-supervised uni-modal approaches. Even with RGB only and depth only as inputs, as shown in the Tables \ref{tab:uni-semi-sup-both}, our framework shows an absolute improvement of up to $2.96\%$ mIoU for RGB and $5.88\%$ mIoU for depth over CPS-Seg\footnote{\label{note:cps-seg}CPS\cite{chen2021-CPS} proposed DeepLabV3+\cite{chen2017rethinking} as base segmentation model, however, we use Segformer \cite{segformer} for a fair comparison and call it CPS-Seg.} \cite{chen2021-CPS}, the current state-of-the-art uni-modal segmentation framework. Thus, M3L effectively uses an additional modality for training to improve the uni-modal segmentation performance label efficiency.

\subsubsection{Linear Fusion}
\label{sec:results-lf}

Table \ref{tab:results-lf} shows the performance of our proposed cross-modal fusion mechanism method, Linear Fusion, compared to other fusion mechanisms when trained with varying amounts of data on both datasets. We find that the simpler Linear Fusion is a strong performer, especially when trained with fewer labels, and gives an improvement of up to $3.6\%$ points mIoU over the state-of-the-art Token Fusion \cite{tokenfusion} by using even fewer parameters. Due to the simplicity and fewer parameters, Linear Fusion also has a faster inference time, computed for a data sample on a single Nvidia A40. These results validate the effectiveness of our proposed Linear Fusion for cross-modal integration.

\subsubsection{Ablation}
\label{sec:results-ablation}
\noindent\textbf{Linear Fusion.}
As mentioned in Section \ref{sec:method-lf}, the fusion weight $\alpha$ is a hyperparameter denoting the linear fusion weight for tokens of the two modalities. Hence, we use the validation set and tune $\alpha$ for the Stanford Indoor dataset \cite{stanfordindoor} when trained with 0.2\% data. As presented in Table \ref{tab:alphatune}, we vary the fusion weight from $0.4$ to $0.9$ with increments of $0.1$ and choose $\alpha=0.8$. We also see that the performance does not vary much with changes in the fusion weight, reflecting that our framework is not sensitive to $\alpha$, as long as it is within a reasonable range.

\noindent\textbf{M3L.} To show the effectiveness of different components of M3L, we present the performance of different components individually on the Stanford Indoor dataset in Table \ref{tab:ablation-M3L}. We first ablate over the simple modality dropout augmentation, which was first proposed in \cite{neverova2015moddrop} for gesture recognition, and then used in \cite{hussen2020modality, 9746613} for other domains like audio-video and animated faces. We use the learned token approach presented in Section \ref{sec:method-M3L} for TF. For URN \cite{urn}, we follow the modality dropout proposed in \cite{urn}. We show that our simple Linear Fusion benefits the most on the MM-Robust metric using just the modality dropout augmentation. Crucially, this augmentation leads to worse performance for the multi-modal input (RGBD) scenario for all three base segmentation models (TF \cite{tokenfusion}, URN \cite{urn}, and Linear Fusion). We then ablate over the use of unlabeled data and show that naively using unlabeled data with or without modality dropout aug does not help with the MM-Robust metric over the sup-only baseline. \emph{M3L}, which uses a \emph{M}ulti-modal teacher for \emph{M}asked \emph{M}odality \emph{L}earning, outperforms all other choices.

%% file: tables/missing-modality-semi-sup-both.tex
\begin{table*}[ht]
\resizebox{\linewidth}{!}{%
\begin{tabular}{rllllllllllll}
\toprule
\multicolumn{1}{c}{\multirow{2}{*}{Method}} & \multicolumn{4}{c}{0.1\%  (49)} & \multicolumn{4}{c}{0.2\% (98)} & \multicolumn{4}{c}{1\% (491)} \\
\multicolumn{1}{c}{} & RGB & Depth & RGBD & \multicolumn{1}{l|}{MM-Robust} & RGB & Depth & RGBD & \multicolumn{1}{c|}{MM-Robust} & RGB & Depth & RGBD & MM-Robust \\ \midrule
Uni-modal\cite{segformer} & 35.43 & 34.05 & ~~~- & \multicolumn{1}{l|}{~~~-} & 39.45 & 35.24 & ~~~- & \multicolumn{1}{l|}{~~~-} & 46.45 & 44.78 & ~~~- & ~~~- \\
TF \cite{tokenfusion} & 29.96 & 29.98 & 40.17 & \multicolumn{1}{l|}{33.37} & 33.11 & 31.47 & 43.04 & \multicolumn{1}{l|}{35.87} & 37.34 & 28.33 & 51.85 & 39.17 \\
URN \cite{urn} & 30.56 & 25.85 & 40.17 & \multicolumn{1}{l|}{32.19} & 35.71 & 25.14 & 45.87 & \multicolumn{1}{l|}{35.57} & 36.25 & 33.27 & 52.07 & 40.53 \\
LF & 33.96 & 25.09 & 42.09 & \multicolumn{1}{l|}{33.71} & 33.66 & 24.61 & 46.60 & \multicolumn{1}{l|}{34.96} & 33.51 & 23.70 & 52.47 & 36.56 \\ \midrule
LF + MT~\cite{mt} & 32.37 & 22.92 & 41.77 & \multicolumn{1}{l|}{32.35} & 33.99 & 23.34 & \underline{48.54} & \multicolumn{1}{l|}{35.29} & 33.65 & 22.42 & 54.32 & 36.80 \\
Ours & \textbf{40.05} \add{7.68}& \textbf{39.93} \add{9.95} & \textbf{44.10} \add{2.01} & \multicolumn{1}{l|}{\textbf{41.36} \add{7.65}} & \textbf{44.62} \add{8.91} & \textbf{42.70} \add{11.23}& \textbf{49.05} \add{0.51} & \multicolumn{1}{l|}{\textbf{45.46} \add{9.59}} & \textbf{49.28} \add{11.94} & \textbf{46.79}\add{13.52} & \textbf{55.48} \add{1.16}& \textbf{50.52} \add{9.99}\\\bottomrule
\multicolumn{13}{c}{\vspace{-0.25cm}}\\
\multicolumn{13}{c}{(a) Stanford Indoor dataset}\\
\multicolumn{13}{c}{\vspace{0.15cm}}\\
\toprule
\multicolumn{1}{c}{\multirow{2}{*}{Method}} & \multicolumn{4}{c}{6.25\%  (297)} & \multicolumn{4}{c}{12.5\% (594)} & \multicolumn{4}{c}{25\% (1189)} \\
\multicolumn{1}{c}{} & RGB & Depth & RGBD & \multicolumn{1}{l|}{MM-Robust} & RGB & Depth & RGBD & \multicolumn{1}{c|}{MM-Robust} & RGB & Depth & RGBD & MM-Robust \\ \midrule
Uni-modal\cite{segformer} & 28.71 & 22.81 & ~~~- & \multicolumn{1}{l|}{~~~-} & 35.33 & 27.6 & ~~~- & \multicolumn{1}{l|}{~~~-} & 38.31 & 30.43 & ~~~- & ~~~- \\
TF \cite{tokenfusion} & 27.97 & 23.58 & 29.31 & \multicolumn{1}{l|}{26.95} & 33.75 & 28.31 & 35.88 & \multicolumn{1}{l|}{32.65} & 37.36 & 31.90 & 39.86 & 36.37 \\
URN \cite{urn}& 28.72 & 12.47 & \underline{31.31} & \multicolumn{1}{l|}{24.17} & 33.66 & 15.62 & 37.62 & \multicolumn{1}{l|}{28.97} & 37.49 & 17.27 & 40.49 & 31.75 \\
LF & \underline{29.69} & 15.75 & \textbf{32.00} & \multicolumn{1}{l|}{25.81} & 35.48 & 17.46 & \underline{39.00} & \multicolumn{1}{l|}{30.65} & 39.15 & 17.66 & \underline{42.09} & 32.97 \\ \midrule
LF + MT~\cite{mt} & \underline{29.57} & 17.86 & 31.11 & \multicolumn{1}{l|}{26.18} & 34.82 & 18.89 & \underline{39.17} & \multicolumn{1}{l|}{30.96} & 38.96 & 21.03 & \underline{41.95} & 33.98 \\ 
Ours & \textbf{29.92} \add{0.23} & \textbf{25.44} \add{1.86} & 30.67 \sub{1.33} & \multicolumn{1}{l|}{\textbf{28.68} \add{1.73}} & \textbf{38.12} \add{2.64}& \textbf{32.29} \add{4.08} & \textbf{39.70} \add{0.53} & \multicolumn{1}{l|}{\textbf{36.70} \add{4.05}} & \textbf{41.31} \add{2.16} & \textbf{34.11} \add{2.21}& \textbf{42.69} \add{0.6} & \textbf{39.37} \add{3.00}\\ \bottomrule
\multicolumn{13}{c}{\vspace{-0.25cm}}\\
\multicolumn{13}{c}{(b) SUN RGBD dataset}
\end{tabular}
}
\caption{\small{\textbf{Missing Modality robustness}. We compare the multi-modal models on three testing scenarios: RGBD, RGB (Depth missing), and Depth (RGB missing). We also report the individual uni-modal model's performance for the two modalities for comparison.}}
\label{tab:missing-modality-semi-sup-both}
\end{table*}

%% file: tables/uni-semi-sup.tex
\begin{table}[t]
\centering
\resizebox{0.85\linewidth}{!}{%
\begin{tabular}{rlll}
\toprule
Method & \multicolumn{1}{l}{0.1\% (49)} & \multicolumn{1}{l}{0.2\% (98)} & \multicolumn{1}{l}{1\% (491)} \\ \midrule
Uni-modal (sup-only) & 35.43 & 39.45 & 46.45 \\
Uni-modal + MT \cite{mt}& 36.59 & 41.50 & 47.04 \\
Uni-modal + CPS-Seg\footnoteref{note:cps-seg} \cite{chen2021-CPS}& 37.09 & 42.75 & 46.37 \\ \midrule
Ours & \textbf{40.05} \add{2.96}& \textbf{44.62} \add{1.87}& \textbf{49.28} \add{2.24}\\ \bottomrule
\multicolumn{4}{c}{\vspace{-0.25cm}}\\
\multicolumn{4}{c}{(a) RGB uni-modal}\\
\multicolumn{4}{c}{}\\
\toprule
Method & \multicolumn{1}{l}{0.1\% (49)} & \multicolumn{1}{l}{0.2\% (98)} & \multicolumn{1}{l}{1\% (491)} \\ \midrule
Uni-modal (sup-only) & 34.05 & 35.24 & 44.78 \\
Uni-modal + MT \cite{mt}& 33.46 & 37.57 & \underline{46.25} \\ 
Uni-modal + CPS-Seg\footnoteref{note:cps-seg} \cite{chen2021-CPS} & 33.56 & 36.71 & 45.71 \\ \midrule
Ours & \textbf{39.93} \add{5.88}& \textbf{42.70} \add{5.13}& \textbf{46.79} \add{0.54}\\ \bottomrule
\multicolumn{4}{c}{\vspace{-0.25cm}}\\
\multicolumn{4}{c}{(b) Depth uni-modal}
\end{tabular}
}
\caption{\small{\textbf{Multi-modal training benefits the uni-modal segmentation results.} Performance with uni-modal input of our proposed LF+M3L method as compared to uni-modal supervised only and semi-supervised (MT, CPS) models on Stanford Indoor dataset \cite{stanfordindoor}}}
\label{tab:uni-semi-sup-both}
\end{table}

%% file: tables/results-lf.tex
\begin{table*}[h]
\centering
\resizebox{\linewidth}{!}{%
\begin{tabular}{rccllll|llll}
\toprule
\multicolumn{1}{c}{\multirow{2}{*}{Method}} & \multicolumn{1}{c}{\multirow{1}{*}{\makecell{Trained\\parameters}}} & \multicolumn{1}{c}{\multirow{1}{*}{\makecell{Inference\\time(ms)}}} & \multicolumn{4}{c|}{Stanford Indoor} & \multicolumn{4}{c}{SUN RGBD} \\
\multicolumn{1}{c}{} &\multicolumn{1}{c}{} &\multicolumn{1}{c}{} & \multicolumn{1}{l}{0.1\% (49)} & \multicolumn{1}{l}{0.2\% (98)} & \multicolumn{1}{l}{1\% (491)} & \multicolumn{1}{l|}{100\% (49199)} & \multicolumn{1}{l}{6.25\% (297)} & \multicolumn{1}{l}{12.5\% (594)} & \multicolumn{1}{l}{25\% (1189)} & \multicolumn{1}{l}{100\% (4757)} \\ \midrule
Uni-modal RGB & 24.73M &17.2& 35.43 & 39.45 & 46.45 & 50.82 & 28.71 & 35.33 & 38.31 & 45.89 \\
Uni-modal Depth & 24.73M & 17.2&34.05 & 35.24 & 44.78 & 52.65 & 22.81 & 27.60 & 30.43 & 36.93 \\
TF \cite{tokenfusion} &26.02M &55.7 & 40.17 & 43.04 & \underline{51.85} & 56.64 & 29.31 & 35.88 & 39.86 & 47.00 \\
URN \cite{urn} &48.93M &36.6 &40.17 & \underline{45.87} & \underline{52.07} & 56.67 & \underline{31.31} & 37.62 & 40.49 & \underline{47.99} \\ \midrule
LF (Ours) & 24.75M&31.7 &\textbf{42.09} \add{1.92} & \textbf{46.60} \add{0.73} & \textbf{52.47} \add{0.4}& \textbf{57.16} \add{0.49}& \textbf{32.00} \add{0.69} & \textbf{39.00} \add{1.38}& \textbf{42.09} \add{1.6} & \textbf{48.17} \add{0.18}\\ \bottomrule
\end{tabular}
}
\vspace{-3mm}
\caption{\small{Comparison of supervised-only multi-modal models trained with varying amounts of data.}}
\label{tab:results-lf}
\end{table*}


%% file: tables/ablation-M3L.tex
\begin{table*}[h]
\resizebox{\linewidth}{!}{%
\begin{tabular}{c|ccc|cccc|cccc|cccc}
\toprule
\multirow{2}{*}{Method} & \multirow{2}{*}{\makecell{Modality\\Dropout}} & \multirow{2}{*}{Unlabeled} & \multirow{2}{*}{M3L} & \multicolumn{4}{c|}{0.1\%  (49)} & \multicolumn{4}{c|}{0.2\% (98)} & \multicolumn{4}{c}{1\% (491)} \\
 &  &  &  & RGB & Depth & RGBD & MM-Robust & RGB & Depth & RGBD & MM-Robust & RGB & Depth & RGBD & MM-Robust \\ \midrule
\multicolumn{1}{r|}{TF \cite{tokenfusion}} &  -& - &  -& 29.96 & 29.98 & 40.17 & 33.37 & 33.11 & 31.47 & 43.04 & 35.87 & 37.34 & 28.33 & 51.85 & 39.17 \\
\multicolumn{1}{r|}{URN \cite{urn}} &  -& - &  -& 30.56 & 25.85 & 40.17 & 32.19 & 35.71 & 25.14 & 45.87 & 35.57 & 36.25 & 33.27 & 52.07 & 40.53 \\
\multicolumn{1}{r|}{TF+MD} & \checkmark &  -& - & 31.1 & 32.34 & 37.23 & 33.56 & 37.79 & 31.95 & 39.9 & 36.55 & 44.43 & 42.7 & 51.17 & 46.1 \\
\multicolumn{1}{r|}{URN+MD} & \checkmark &  -&  -& 34.6 & 33.04 & 39.82 & 35.82 & 39.25 & 36.19 & 43.78 & 39.74 & 45.91 & 44.02 & 52.6 & 47.51 \\ \midrule
\multirow{5}{*}{Ours} &  -& - &  -& 33.96 & 25.09 & 42.09 & 33.71 & 33.11 & 24.61 & 46.6 & 34.96 & 33.51 & 23.7 & 52.47 & 36.56 \\
 & \checkmark &  -& - & 36.26 & 33.79 & 41.41 & 37.15 & 41.06 & 36.13 & 45.53 & 40.91 & 47.14 & 44.86 & 53.18 & 48.39 \\
 &  -& \checkmark &  -& 32.37 & 22.92 & 41.77 & 32.35 & 33.99 & 23.34 & \underline{48.54} & 35.29 & 22.42 & 33.65 & 54.32 & 36.8 \\
 & \checkmark & \checkmark &  -& 35.11 & 34.69 & 39.06 & 36.29 & 41.78 & 39.52 & 46.81 & 42.7 & 46.12 & \textbf{48.22} & 53.97 & 49.44 \\
 & \checkmark & \checkmark & \checkmark & \textbf{40.05} & \textbf{39.93} & \textbf{44.1} & \textbf{41.36} & \textbf{44.62} & \textbf{42.7} & \textbf{49.05} & \textbf{45.46} & \textbf{49.28} & 46.79 & \textbf{55.48} & \textbf{50.52} \\ \bottomrule
\end{tabular}
}
\vspace{-3mm}
\caption{\small{Ablation study for our proposed approaches Linear Fusion and M3L. \emph{Our}'s use Linear Fusion as the base segmentation model.}}
\label{tab:ablation-M3L}
\end{table*}

%% file: tables/alphatune.tex
\begin{table}[t]
\centering
\resizebox{0.9\linewidth}{!}{
\begin{tabular}{ccccccc}
\toprule
Fusion weight ($\alpha$) & 0.4 & 0.5 & 0.6 & 0.7 & 0.8 & 0.9 \\ \midrule
Val mIoU & 56.39 & 57.30 & 57.16 & 56.94 & \textbf{57.81} & 56.30\\\bottomrule
\end{tabular}
}
\caption{\small{Validation mean IoU to tune fusion weight on Stanford Indoor dataset with 0.2\% labels.}}
\vspace{-3mm}
\label{tab:alphatune}
\end{table}

%% file: sections/conclusion.tex
\section{Conclusion}
We explore a new problem of semi-supervised multi-modal semantic segmentation and address its two major challenges: limited supervision during training and missing modalities during testing. To tackle these challenges, we propose (a) Linear Fusion, a simple yet effective fusion mechanism that achieves state-of-the-art results with limited supervision, and (b) M3L, a semi-supervised framework that makes the models robust to a realistic scenario of missing modalities and keeps performance better than its uni-modal counterparts even if multiple modalities are not guaranteed at test time. We build a new semi-supervised multi-modal semantic segmentation benchmark and show the effectiveness of our proposed methods against competitive state-of-art.

%% file: sections/appendix.tex
\section{Appendix}
\input{sections/appendix-subsections/appendix-metrics}
\input{sections/appendix-subsections/appendix-examples}
\input{sections/appendix-subsections/appendix-implementation}

%% file: sections/appendix-subsections/appendix-metrics.tex
\subsection{Additional Metrics}
We reported the mean IoU metrics in Tables \ref{tab:missing-modality-semi-sup-both} and \ref{tab:uni-semi-sup-both}. Here, we also report the other two popular metrics used in segmentation, namely, mean accuracy (mAcc.) and pixel accuracy (Pix. Acc.). Mean accuracy is the average  classification accuracy of a class whereas pixel accuracy is macro classification accuracy for all pixels. Table \ref{tab:appendix-stanfordindoor} reports the performance for Stanford Indoor \cite{stanfordindoor} dataset. Table \ref{tab:appendix-sunrgbd} reports the performance for SUN RGBD \cite{sunrgbd}. As seen, on the MM-Robust, which measures the average performance across three testing scenarios, our method outperforms all baselines for all three metrics.  

We also report these three metrics for the uni-modal semi-supervised results in Table \ref{tab:uni-modal}. We can see that even when tested with a single modality, our method performs better than state-of-the-art uni-modal semi-supervised methods on all three metrics. Since CPS \cite{chen2021-CPS} was proposed originally with the DeepLabV3+~\cite{chen2017rethinking} base segmentation model, we also compare our model with CPS-Dlv3p with ResNet-101 encoder.

\input{tables/appendix/semi-sup-stanford}
\input{tables/appendix/semi-sup-sunrgbd}
\input{tables/appendix/uni-modal}

%% file: tables/appendix/semi-sup-stanford.tex
\begin{table*}[h]
\centering
\resizebox{0.96\linewidth}{!}{
\begin{tabular}{rccc|ccc|ccc|ccc}
\toprule
\multicolumn{1}{c}{\multirow{2}{*}{Method}} & \multicolumn{3}{c|}{RGB} & \multicolumn{3}{c|}{Depth} & \multicolumn{3}{c|}{RGBD} & \multicolumn{3}{c}{MM-Robust} \\
\multicolumn{1}{c}{} & mIoU & mAcc. & Pix. Acc. & mIoU & mAcc. & Pix. Acc. & mIoU & mAcc. & Pix. Acc. & mIoU & mAcc. & Pix. Acc. \\ \midrule
Uni-modal RGB & 35.43 & 47.79 & 63.93 &-  & - & - & - & - & - & - &  &  \\
Uni-modal Depth & - & - & - & 34.05 & 45.63 & 62.56 & - & - & - & - & - &  \\
TF \cite{tokenfusion} & 29.96 & 42.63 & 58.41 & 29.98 & 41.36 & 58.82 & 40.17 & 50.86 & 68.82 & 33.37 & 44.95 & 62.02 \\
URN \cite{urn} & 30.56 & 44.30 & 57.82 & 25.85 & 37.16 & 56.07 & 40.17 & 52.75 & 67.23 & 32.19 & 44.74 & 60.37 \\
LF & 33.96 & 47.86 & 60.06 & 25.09 & 36.93 & 49.93 & 42.09 & \textbf{55.25} & 69.23 & 33.71 & 46.68 & 59.74 \\\midrule
LF + MT & 32.37 & 42.23 & 59.91 & 22.92 & 30.13 & 56.18 & 41.77 & 52.08 & 68.22 & 32.35 & 41.48 & 61.44 \\
LF + M3L & \textbf{40.05} & \textbf{50.47} & \textbf{69.09} & \textbf{39.93} & \textbf{49.97} & \textbf{70.91} & \textbf{44.10} & 53.79 & \textbf{72.94} & \textbf{41.36} & \textbf{51.41} & \textbf{70.98} \\ \bottomrule

\multicolumn{13}{c}{\vspace{0.1cm}}\\
\multicolumn{13}{c}{(a) 0.1\% (49) labeled data}\\
\multicolumn{13}{c}{\vspace{0.4cm}}\\

\toprule
\multicolumn{1}{c}{\multirow{2}{*}{Method}} & \multicolumn{3}{c|}{RGB} & \multicolumn{3}{c|}{Depth} & \multicolumn{3}{c|}{RGBD} & \multicolumn{3}{c}{MM-Robust} \\
\multicolumn{1}{c}{} & mIoU & mAcc. & Pix. Acc. & mIoU & mAcc. & Pix. Acc. & mIoU & mAcc. & Pix. Acc. & mIoU & mAcc. & Pix. Acc. \\ \midrule
Uni-modal RGB & 39.45 & 49.97 & 65.95 & - & - & - & - & - & - & - & - & - \\
Uni-modal Depth & - & - & - & 35.24 & 46.97 & 64.10 & - & - & - & - & - & - \\
TF \cite{tokenfusion} & 33.11 & 44.25 & 60.92 & 31.47 & 42.55 & 59.27 & 43.04 & 52.35 & 70.33 & 35.87 & 46.38 & 63.51 \\
URN \cite{urn} & 35.71 & 46.74 & 62.20 & 25.14 & 37.42 & 56.87 & 45.87 & 56.20 & 70.90 & 35.57 & 46.79 & 63.32 \\
LF & 33.51 & 41.9 & 60.47 & 23.7 & 30.75 & 54.06 & 46.6 & \underline{57.37} & 71.87 & 36.56 & 45 & 63.74 \\ \midrule
LF + MT & 33.65 & 42.58 & 60.92 & 22.42 & 29.04 & 52.71 & \underline{48.54} & \underline{57.67} & 74.85 & 36.8 & 45.52 & 63.77 \\
LF + M3L & \textbf{44.62} & \textbf{54.99} & \textbf{71.28} & \textbf{42.70} & \textbf{52.60} & \textbf{71.91} & \textbf{49.05} & \textbf{58.28} & \textbf{75.01} & \textbf{45.46} & \textbf{55.29} & \textbf{72.73} \\ \bottomrule

\multicolumn{13}{c}{\vspace{0.1cm}}\\
\multicolumn{13}{c}{(a) 0.2\% (98) labeled data}\\
\multicolumn{13}{c}{\vspace{0.4cm}}\\

\toprule
\multicolumn{1}{c}{\multirow{2}{*}{Method}} & \multicolumn{3}{c|}{RGB} & \multicolumn{3}{c|}{Depth} & \multicolumn{3}{c|}{RGBD} & \multicolumn{3}{c}{MM-Robust} \\
\multicolumn{1}{c}{} & mIoU & mAcc. & Pix. Acc. & mIoU & mAcc. & Pix. Acc. & mIoU & mAcc. & Pix. Acc. & mIoU & mAcc. & Pix. Acc. \\ \midrule
Uni-modal RGB & 46.45 & 56.2 & 71.73 & - & - & - & - & - & - & - & - & - \\
Uni-modal Depth & - & - & - & 44.78 & 55.24 & 72.40 & - & - & - & - & - & - \\
TF \cite{tokenfusion}& 37.34 & 45.83 & 65.86 & 28.33 & 41.07 & 57.29 & 51.85 & 62.30 & 75.82 & 39.17 & 49.73 & 66.32 \\
URN \cite{urn}& 36.25 & 45.35 & 64.35 & 33.27 & 45.11 & 62.39 & 52.07 & 61.04 & 76.69 & 40.53 & 50.5 & 67.81 \\
LF & 33.51 & 41.90 & 60.47 & 23.70 & 30.75 & 54.06 & 52.47 & 62.34 & 76.69 & 36.56 & 45.00 & 63.74 \\ \midrule
LF + MT & 33.65 & 42.58 & 60.92 & 22.42 & 29.04 & 52.71 & \underline{54.32} & \textbf{64.93} & \underline{77.69} & 36.80 & 45.52 & 63.77 \\
LF + M3L & \textbf{49.28} & \textbf{59.03} & \textbf{73.86} & \textbf{46.79} & \textbf{57.41} & \textbf{74.11} & \textbf{55.48} & \underline{64.78} & \textbf{78.59} & \textbf{50.52} & \textbf{60.41} & \textbf{75.52} \\ \bottomrule

\multicolumn{13}{c}{\vspace{0.1cm}}\\
\multicolumn{13}{c}{(a) 1\% (491) labeled data}\\

\end{tabular}
}
\caption{We compare the multi-modal models on three testing scenarios: RGBD, RGB (Depth missing),
and Depth (RGB missing) using three metrics on Stanford Indoor dataset. We also report the individual uni-modal model’s performance for the two modalities for comparison.}
\label{tab:appendix-stanfordindoor}
\end{table*}

%% file: tables/appendix/semi-sup-sunrgbd.tex
\begin{table*}[h]
\centering
\resizebox{0.96\linewidth}{!}{
\begin{tabular}{rccc|ccc|ccc|ccc}
\toprule
\multicolumn{1}{c}{\multirow{2}{*}{Method}} & \multicolumn{3}{c|}{RGB} & \multicolumn{3}{c|}{Depth} & \multicolumn{3}{c|}{RGBD} & \multicolumn{3}{c}{MM-Robust} \\
\multicolumn{1}{c}{} & mIoU & mAcc. & Pix. Acc. & mIoU & mAcc. & Pix. Acc. & mIoU & mAcc. & Pix. Acc. & mIoU & mAcc. & Pix. Acc. \\ \midrule
Uni-modal RGB & 28.71 & 37.21 & 73.36 & - & - & - & - & - & - & - & - & - \\
Uni-modal Depth & - & - & - & 22.81 & 29.81 & 70.38 & - & - & - & - & - & - \\
TF \cite{tokenfusion} & 27.97 & 36.05 & 72.15 & 23.58 & 30.73 & 70.62 & 29.31 & 35.93 & 74.82 & 26.95 & 34.24 & 72.53 \\
URN \cite{urn}& 28.72 & \textbf{39.60} & 72.30 & 12.47 & 18.00 & 61.11 & 31.31 & 40.54 & 74.93 & 24.17 & 32.71 & 69.45 \\
LF & 29.69 & \underline{39.17} & 73.83 & 15.75 & 22.24 & 64.81 & \textbf{32.00} & \textbf{41.48} & \underline{75.92} & 25.81 & 34.30 &71.52 \\ \midrule
LF + MT & \underline{29.57} & 37.32 & 74.42 & 17.86 & 23.10 & 67.16 & \underline{31.11} & 38.76 & 76.12 & 26.18 & 33.06 & 72.57 \\
LF + M3L & \textbf{29.92} & 36.83 & \textbf{75.19} & \textbf{25.44} & \textbf{32.30} & \textbf{72.32} & 30.67 & 37.20 & \textbf{76.36} & \textbf{28.68} & \textbf{35.44} & \textbf{74.62} \\ \bottomrule

\multicolumn{13}{c}{\vspace{0.1cm}}\\
\multicolumn{13}{c}{(a) 6.25\% (297) labeled data}\\
\multicolumn{13}{c}{\vspace{0.4cm}}\\

\toprule
\multicolumn{1}{c}{\multirow{2}{*}{Method}} & \multicolumn{3}{c|}{RGB} & \multicolumn{3}{c|}{Depth} & \multicolumn{3}{c|}{RGBD} & \multicolumn{3}{c}{MM-Robust} \\
\multicolumn{1}{c}{} & mIoU & mAcc. & Pix. Acc. & mIoU & mAcc. & Pix. Acc. & mIoU & mAcc. & Pix. Acc. & mIoU & mAcc. & Pix. Acc. \\ \midrule
Uni-modal RGB & 35.33 & 45.2 & 76.15 & - & - & - & - & - & - & - & - & - \\
Uni-modal Depth & - & - & - & 27.60 & 35.54 & 72.49 & - & - & - & - & - & - \\
TF \cite{tokenfusion}& 33.75 & 43.99 & 74.42 & 28.31 & 37.04 & 72.14 & 35.88 & 43.93 & 76.96 & 32.65 & 41.65 & \underline{74.51} \\
URN \cite{urn}& 33.66 & 45.67 & 74.14 & 15.62 & 21.68 & 63.74 & 37.62 & 47.55 & 77.41 & 28.97 & 38.30 & 71.76 \\
LF & 35.48 & 46.32 & 75.75 & 17.46 & 24.29 & 65.04 & \underline{39.00} & \textbf{49.13} & \underline{78.20} & 30.65 & 39.91 & 73.00 \\ \midrule
LF + MT & 34.82 & 45.55 & 75.57 & 18.89 & 28.33 & 66.75 & \underline{39.17} & 47.70 & \underline{79.02} & 30.96 & 40.53 & 73.78 \\
LF + M3L & \textbf{38.12} & \textbf{46.93} & \textbf{77.80} & \textbf{32.29} & \textbf{40.96} & \textbf{74.91} & \textbf{39.70} & 47.97 & \textbf{79.05} & \textbf{36.70} & \textbf{45.29} & \textbf{77.25} \\ \bottomrule

\multicolumn{13}{c}{\vspace{0.1cm}}\\
\multicolumn{13}{c}{(a) 12.5\% (594) labeled data}\\
\multicolumn{13}{c}{\vspace{0.4cm}}\\

\toprule
\multicolumn{1}{c}{\multirow{2}{*}{Method}} & \multicolumn{3}{c|}{RGB} & \multicolumn{3}{c|}{Depth} & \multicolumn{3}{c|}{RGBD} & \multicolumn{3}{c}{MM-Robust} \\
\multicolumn{1}{c}{} & mIoU & mAcc. & Pix. Acc. & mIoU & mAcc. & Pix. Acc. & mIoU & mAcc. & Pix. Acc. & mIoU & mAcc. & Pix. Acc. \\ \midrule
Uni-modal RGB & 38.31 & 48.30 & 77.66 & - & - & - & - & - & - & - & - & - \\
Uni-modal Depth & - & - & - & 30.43 & 38.53 & 73.70 & - & - & - & - & - & - \\
TF \cite{tokenfusion}& 37.36 & 48.23 & 76.15 & 31.90 & 40.50 & 73.79 & 39.86 & 48.26 & 78.67 & 36.37 & 45.66 & 76.20 \\
URN \cite{urn}& 37.49 & 49.20 & 76.24 & 17.27 & 22.12 & 64.68 & 40.49 & 50.7 & 78.87 & 31.75 & 40.67 & 73.26 \\
LF & 39.15 & 50.27 & 77.53 & 17.66 & 25.64 & 66.67 & \underline{42.09} & \textbf{52.32} & 79.78 & 32.97 & 42.74 & 74.66 \\ \midrule
LF + MT & 38.96 & 49.47 & 77.38 & 21.03 & 27.71 & 68.81 & \underline{41.95} & \underline{51.99} & 79.57 & 33.98 & 43.06 & 75.25 \\
LF + M3L & \textbf{41.31} & \textbf{51.01} & \textbf{79.15} & \textbf{34.11} & \textbf{42.91} & \textbf{75.58} & \textbf{42.69} & \underline{52.03} & \textbf{80.4} & \textbf{39.37} & \textbf{48.65} & \textbf{78.38} \\ \bottomrule

\multicolumn{13}{c}{\vspace{0.1cm}}\\
\multicolumn{13}{c}{(a) 25\% (1189) labeled data}\\

\end{tabular}
}
\caption{We compare the multi-modal models on three testing scenarios: RGBD, RGB (Depth missing),
and Depth (RGB missing) using three metrics on SUN RGBD dataset. We also report the individual uni-modal model’s performance for the two modalities for comparison.}
\label{tab:appendix-sunrgbd}
\end{table*}

%% file: tables/appendix/uni-modal.tex
\begin{table*}[]
\centering
\resizebox{0.9\linewidth}{!}{
\begin{tabular}{rccccccccc}
\toprule
\multirow{2}{*}{Method} & \multicolumn{3}{c}{0.1 \% (49)} & \multicolumn{3}{c}{0.2\% (98)} & \multicolumn{3}{c}{1\% (491)} \\
 & mIoU & mAcc. & \multicolumn{1}{c|}{Pix. Acc.} & mIoU & mAcc. & \multicolumn{1}{c|}{Pix. Acc.} & mIoU & mAcc. & Pix. Acc. \\ \midrule
Uni-modal (sup only) & 35.43 & 47.79 & \multicolumn{1}{c|}{63.93} & 39.45 & 49.97 & \multicolumn{1}{c|}{65.95} & 46.45 & 56.2 & 71.73 \\
Uni-modal + MT \cite{mt} & 36.59 & 46.18 & \multicolumn{1}{c|}{65.43} & 41.5 & 52.78 & \multicolumn{1}{c|}{69.16} & 47.04 & 56.96 & 72.6 \\
Uni-modal + CPS-Dlv3p \cite{chen2021-CPS}& 33.09 & 42.12 & \multicolumn{1}{c|}{62.56} & 37.95 & 48.16 & \multicolumn{1}{c|}{65.8} & 44.22 & 53.85 & 70.81 \\
Uni-modal + CPS-Seg\footnoteref{note:cps-seg} \cite{chen2021-CPS} & 37.09 & 48.41 & \multicolumn{1}{c|}{65.97} & 42.75 & 51.61 & \multicolumn{1}{c|}{69.96} & 46.37 & 56.32 & 72.86 \\ \midrule
Ours & \textbf{40.05} & \textbf{50.47} & \multicolumn{1}{c|}{\textbf{69.09}} & \textbf{44.62} & \textbf{54.99} & \multicolumn{1}{c|}{\textbf{71.28}} & \textbf{49.28} & \textbf{59.03} & \textbf{73.86} \\ \bottomrule

\multicolumn{10}{c}{\vspace{0.1cm}}\\
\multicolumn{10}{c}{(a) RGB uni-modal}\\
\multicolumn{10}{c}{\vspace{0.4cm}}\\

\toprule
\multirow{2}{*}{Method} & \multicolumn{3}{c}{0.1 \% (49)} & \multicolumn{3}{c}{0.2\% (98)} & \multicolumn{3}{c}{1\% (491)} \\
 & mIoU & mAcc. & \multicolumn{1}{c|}{Pix. Acc.} & mIoU & mAcc. & \multicolumn{1}{c|}{Pix. Acc.} & mIoU & mAcc. & Pix. Acc. \\ \midrule
Uni-modal (sup only) & 34.05 & 45.63 & \multicolumn{1}{c|}{62.56} & 35.24 & 46.97 & \multicolumn{1}{c|}{64.1} & 44.78 & 55.24 & 72.4 \\
Uni-modal + MT \cite{mt} & 33.46 & 43.18 & \multicolumn{1}{c|}{62.45} & 37.57 & 48.51 & \multicolumn{1}{c|}{67.5} & 46.25 & 56.01 & 74.68 \\
Uni-modal + CPS-Dlv3p \cite{chen2021-CPS} & 33.43 & 43.05 & \multicolumn{1}{c|}{64.97} & 35.93 & 47.52	&\multicolumn{1}{c|}{64.11} & 45.50 & 55.13 & 74.04 \\
Uni-modal + CPS-Seg\footnoteref{note:cps-seg} \cite{chen2021-CPS}& 33.56 & 42.67 & \multicolumn{1}{c|}{65.75} & 36.71 & 47.05 & \multicolumn{1}{c|}{66.64} & 45.71 & 55.4 & 74.32 \\ \midrule
Ours & \textbf{39.93} & \textbf{49.97} & \multicolumn{1}{c|}{\textbf{70.91}} & \textbf{42.7} & \textbf{52.6} & \multicolumn{1}{c|}{\textbf{71.91}} & \textbf{46.79} & \textbf{57.41} & \textbf{74.11} \\ \bottomrule

\multicolumn{10}{c}{\vspace{0.1cm}}\\
\multicolumn{10}{c}{(a) Depth uni-modal}\\

\end{tabular}
}
\caption{Uni-modal semi-supervised segmentation. LF when trained with M3L (ours) beats state-of-the-art uni-modal semi-supervised frameworks when tested with a single modality (RGB (a) or Depth (b) modality) as input.}
\label{tab:uni-modal}
\end{table*}

%% file: sections/appendix-subsections/appendix-examples.tex
\subsection{Qualitative Examples}
We also show qualitative results for randomly chosen examples images from the Stanford Indoor \cite{stanfordindoor} dataset. 
Figure \ref{fig:example} compares different base segmentation multi-modal models with Linear Fusion and the proposed M3L semi-supervised framework with supervised-only and mean teacher \cite{mt} frameworks when trained using only $0.1\% ~(49)$ labels. 
We can see that when our model is trained with M3L, the segmentation performance is superior to other supervised or semi-supervised baselines.

We also visualize how the segmentation is affected when a modality is missing. In Figures \ref{fig:example2}, we see that when the missing modality robustness is left untreated (when trained with mean teacher \cite{mt}), the performance is sensitive to the presence of both modalities. In the realistic scenario of missing modalities, the performance degrades significantly. However, when the model is trained with our proposed M3L framework, the predictions can hold up the quality even with missing modalities. 
\input{figures/example3}
\input{figures/example4}
\input{figures/example}
\input{figures/example2}

In Figures \ref{fig:example3}, \ref{fig:example4}, we see an example where the depth modality plays an more important role as the image captures the inside of a room through a door. This information is represented well in the depth modality. If the missing modality problem is left untreated as in the mean teacher \cite{mt} framework, when depth is missing during inference, the prediction worsens significantly as seen in Figure \ref{fig:example3}. However, when treated properly using the proposed M3L framework, even with missing depth, the performance holds up as shown in Figure \ref{fig:example4}.

%% file: figures/example3.tex
\begin{figure}[]
\begin{center}
\includegraphics[width=0.9\linewidth]{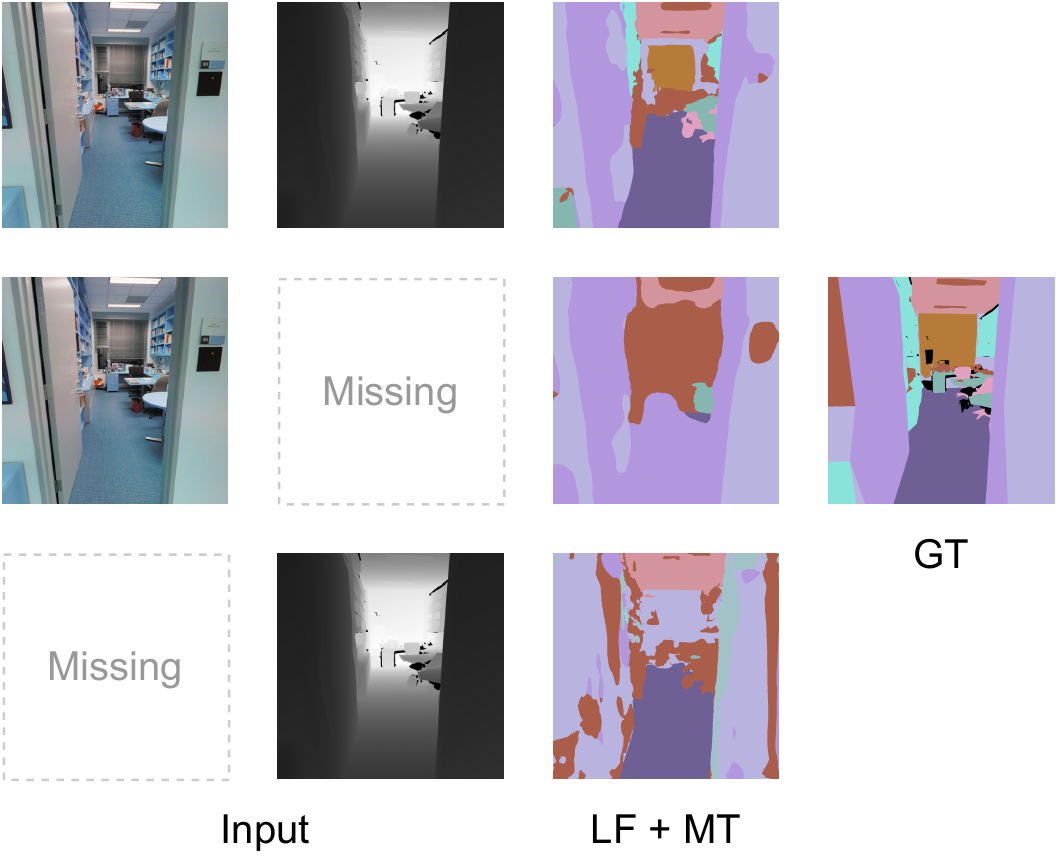}
\end{center}
\caption{\small{An example to show that when Linear Fusion (LF) is trained with mean teacher (MT) \cite{mt}, it is sensitive to the presence of both modalities.}}
\label{fig:example3}
\end{figure}

%% file: figures/example4.tex
\begin{figure}
\begin{center}
\includegraphics[width=0.9\linewidth]{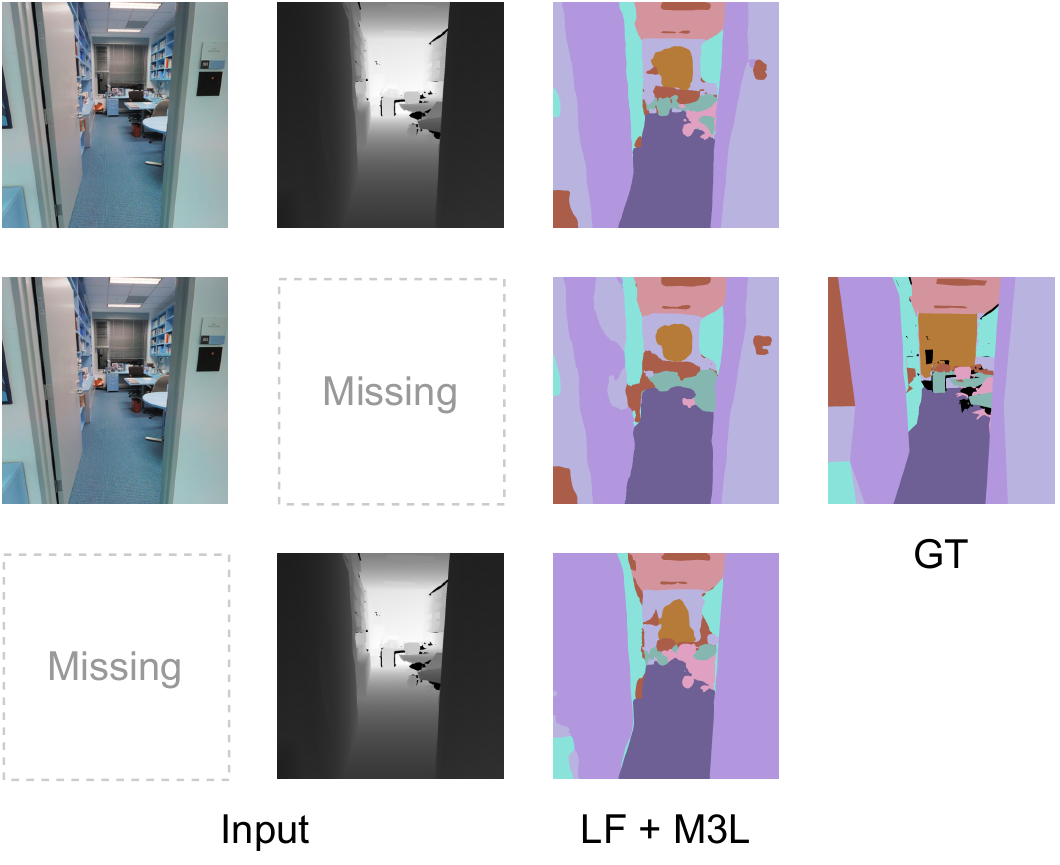}
\end{center}
\caption{\small{When Linear Fusion is trained with our proposed M3L framework, the predictions are robust to the missing modalities.}}
\label{fig:example4}
\end{figure}

%% file: figures/example.tex
\begin{figure*}
\begin{center}
\includegraphics[width=\linewidth]{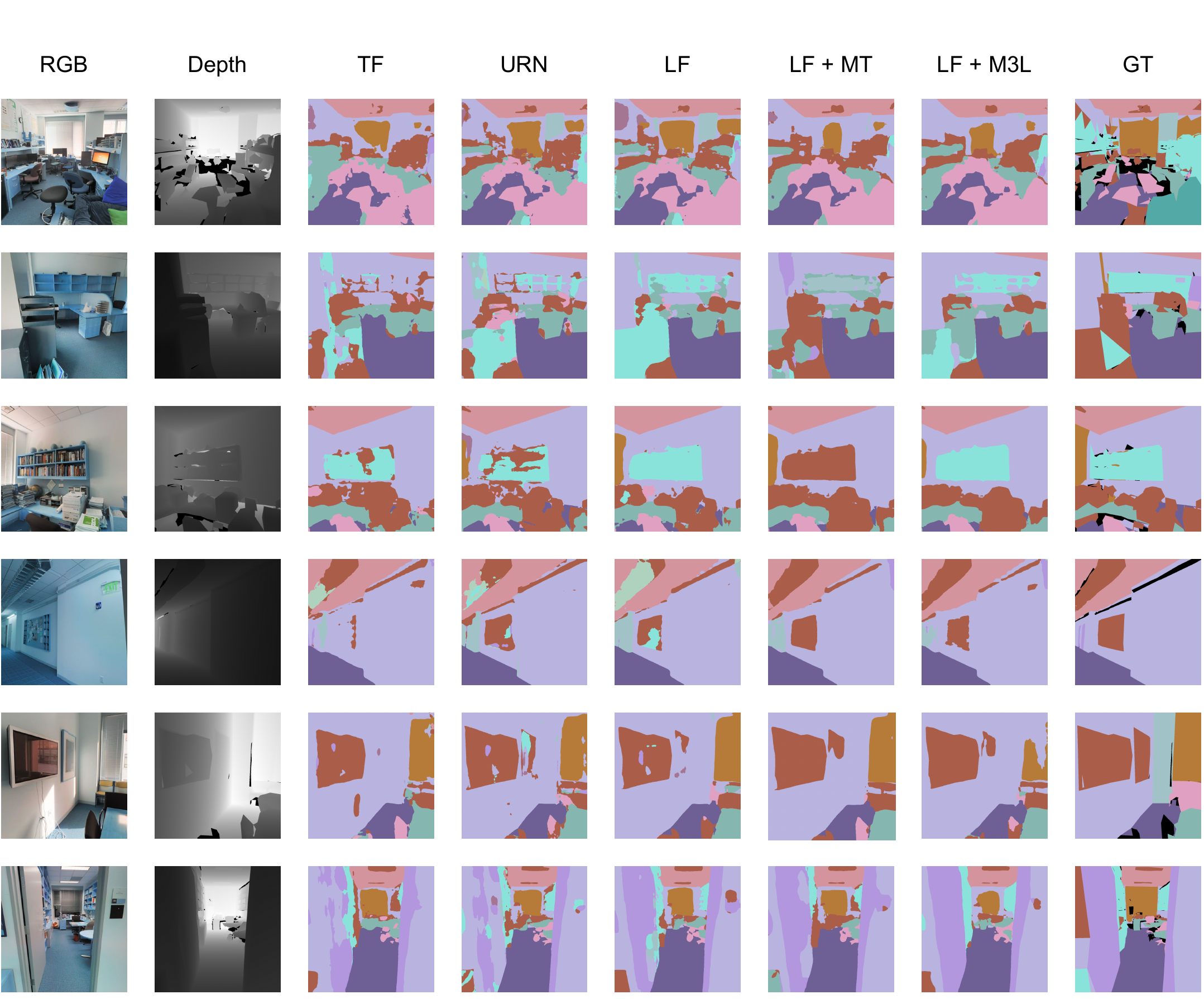}
\end{center}
\caption{Examples for different multi-modal models trained with supervised and semi-supervised frameworks.}
\label{fig:example}
\end{figure*}

%% file: figures/example2.tex
\begin{figure*}
\begin{center}
\includegraphics[width=\linewidth]{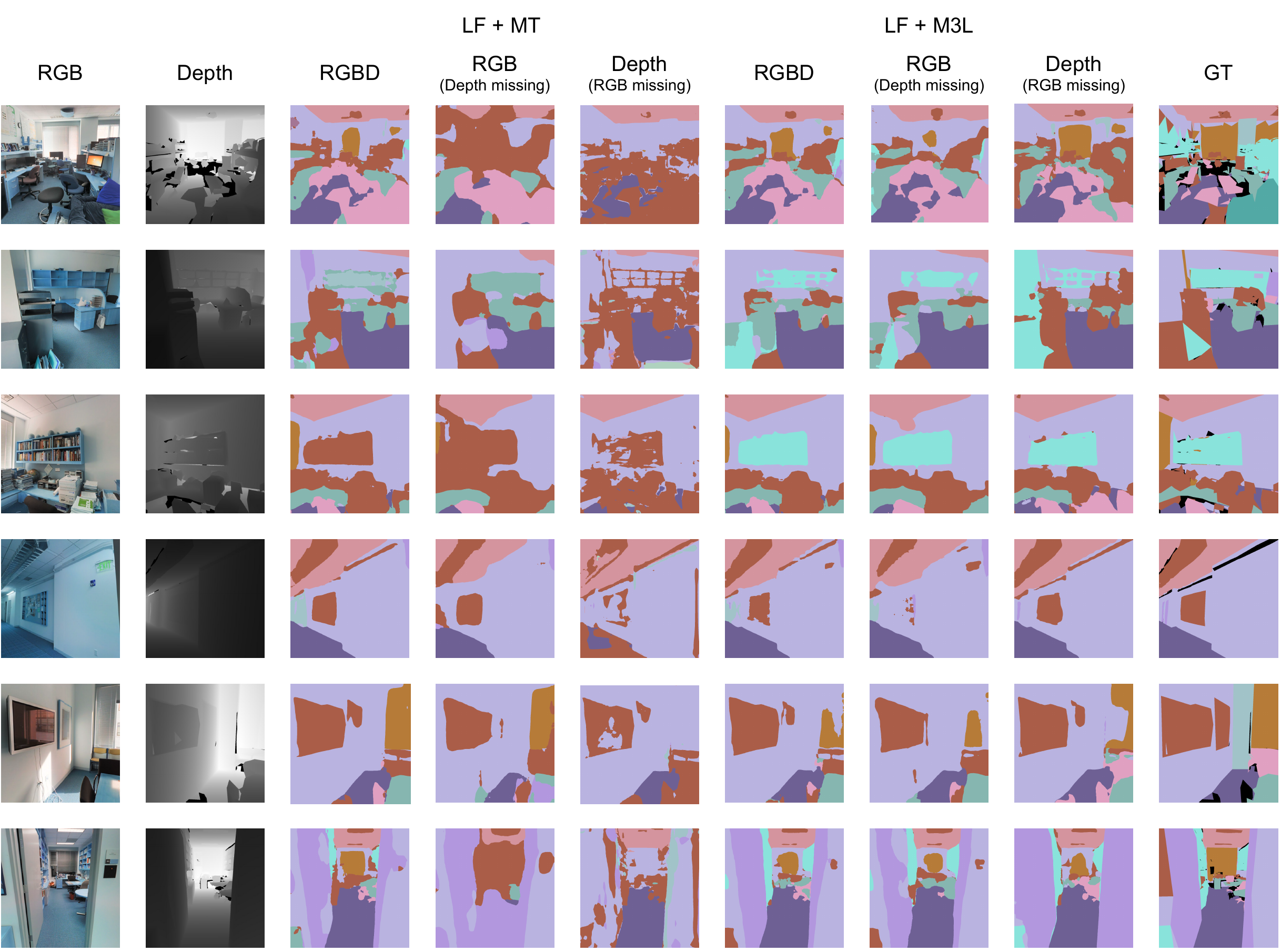}
\end{center}
\caption{Examples for visualizing drop in performance when a modality is missing and robustness to missing modality when trained with the propsed M3L framework.}
\label{fig:example2}
\end{figure*}

%% file: sections/appendix-subsections/appendix-implementation.tex
\subsection{Additional Implementation Details}
To train the proposed segmentation model, Linear Fusion with the proposed semi-supervised training framework M3L, we use a batch size of 16 and load 16 labeled and 16 unlabeled data samples in a batch. We calculate the supervised loss on the 16 labeled samples. Since the unsupervised loss is calculated on both the labeled and unlabeled samples and requires a different forward pass on the labeled samples, we make a copy of the labeled samples and compute the unsupervised loss on this copy and the unlabeled samples. Thus, we pass a batch of 48 instances to the model with 16 labeled, 16 labeled (but same examples) and 16 unlabeled, where the masking is done randomly in the last 32 samples of the batch. The ground truth is a single batch of 16 samples (corresponding to the first 16 samples in feed forward). 
For modality masking in the student input, we randomly choose either RGB or Depth or None modality to mask. As mentioned, we use the multi-class cross entropy loss for the unsupervised loss and use the OHEM loss \cite{ohem} as supervised loss with a threshold of $0.7$. We ignore the supervised loss for pixels with ground truth class missing. We train our model for 5 epochs (one epoch is defined as passing over all training data, and not just the labeled data, once) for Stanford Indoor dataset and 50 for SUN RGBD dataset which results in $15300$ iterations and $14700$ iterations respectively for both the datasets, irrespective of the labeled and unlabeled ratio.